\documentclass[conference]{IEEEtran}
\IEEEoverridecommandlockouts
\usepackage{cite}
\usepackage{amsmath,amssymb,amsfonts}
\usepackage{algorithmic}
\usepackage{graphicx}
\usepackage{textcomp}
\usepackage{xcolor}
\usepackage{pifont}  
\usepackage{bbding}
\usepackage{overpic}
\usepackage{subcaption}
\usepackage{booktabs}

\usepackage{multirow}
\newcommand{\cmark}{\ding{51}}
\newcommand{\xmark}{\ding{55}}
\newcommand{\Tref}[1]{Table~\ref{#1}}

\newcommand{\fref}[1]{Fig.~\ref{#1}}

\newcommand{\nero}{NeRO~\cite{liu2023nero}}
\newcommand{\neus}{NeuS~\cite{wang2021neus}}
\newcommand{\supernormal}{SN~\cite{cao2024supernormal}}
\newcommand{\rnbneus}{Rnb-NeuS~\cite{brument2024rnb}}
\newcommand{\vminer}{VMINer~\cite{vminer}}
\def\BibTeX{{\rm B\kern-.05em{\sc i\kern-.025em b}\kern-.08em
    T\kern-.1667em\lowezr.7ex\hbox{E}\kern-.125emX}}
\begin{document}

\title{RotatedMVPS: Multi-view Photometric Stereo with Rotated 
 Natural Light}

    \author{Songyun Yang$^{1\dagger}$, Yufei Han$^{1\dagger}$, Jilong Zhang$^1$, Kongming Liang$^1$, Peng Yu$^1$, Zhaowei Qu$^{1*}$, Heng Guo$^1$\\
\thanks{$^{\dagger}$ Equal contribution. $^{*}$ Corresponding author.}
$^{1}$Beijing University of Posts and Telecommunications, China\\
{\tt\small \{songyunyang, hanyufei, jilongzhang, liangkongming, yupeng, zwqu, guoheng\}@bupt.edu.cn}
}

\maketitle
\begin{abstract}
          Multiview photometric stereo (MVPS) seeks to recover high-fidelity surface shapes and reflectances from images captured under varying views and illuminations. However, existing MVPS methods often require controlled darkroom settings for varying illuminations or overlook the recovery of reflectances and illuminations properties, limiting their applicability in natural illumination scenarios and downstream inverse rendering tasks. In this paper, we propose RotatedMVPS to solve shape and reflectance recovery under rotated natural light, achievable with a practical rotation stage. By ensuring light consistency across different camera and object poses, our method reduces the unknowns associated with complex environment light. Furthermore, we integrate data priors from off-the-shelf learning-based single-view photometric stereo methods into our MVPS framework, significantly enhancing the accuracy of shape and reflectance recovery. Experimental results on both synthetic and real-world datasets demonstrate the effectiveness of our approach.
          \end{abstract}
\section{Introduction}
Multiview photometric stereo (MVPS) aims to recover high-fidelity surface shapes and reflectances from images captured under varying views and illuminations. Traditional MVPS~\cite{yang2022ps, zhao2023mvpsnet,kaya2022uncertainty,kaya2022neural} methods rely on controlled darkroom environments with directional light to ensure consistent and well-defined light conditions. However, they are impractical for many real-world applications. Therefore, extending MVPS under natural environment illumination remains an open problem.

To address natural environment light in MVPS, recent methods such as SuperNormal~\cite{cao2024supernormal} and Rnb-NeuS~\cite{brument2024rnb} leverage learning-based photometric stereo techniques for natural light settings. These approaches initially estimate per-view surface normals and integrate them to recover shapes. However, they fall short for inverse rendering tasks because they fail to estimate environment light and reflectance properties. 

\begin{figure}
          \centering
          \begin{overpic}[width=1\linewidth]{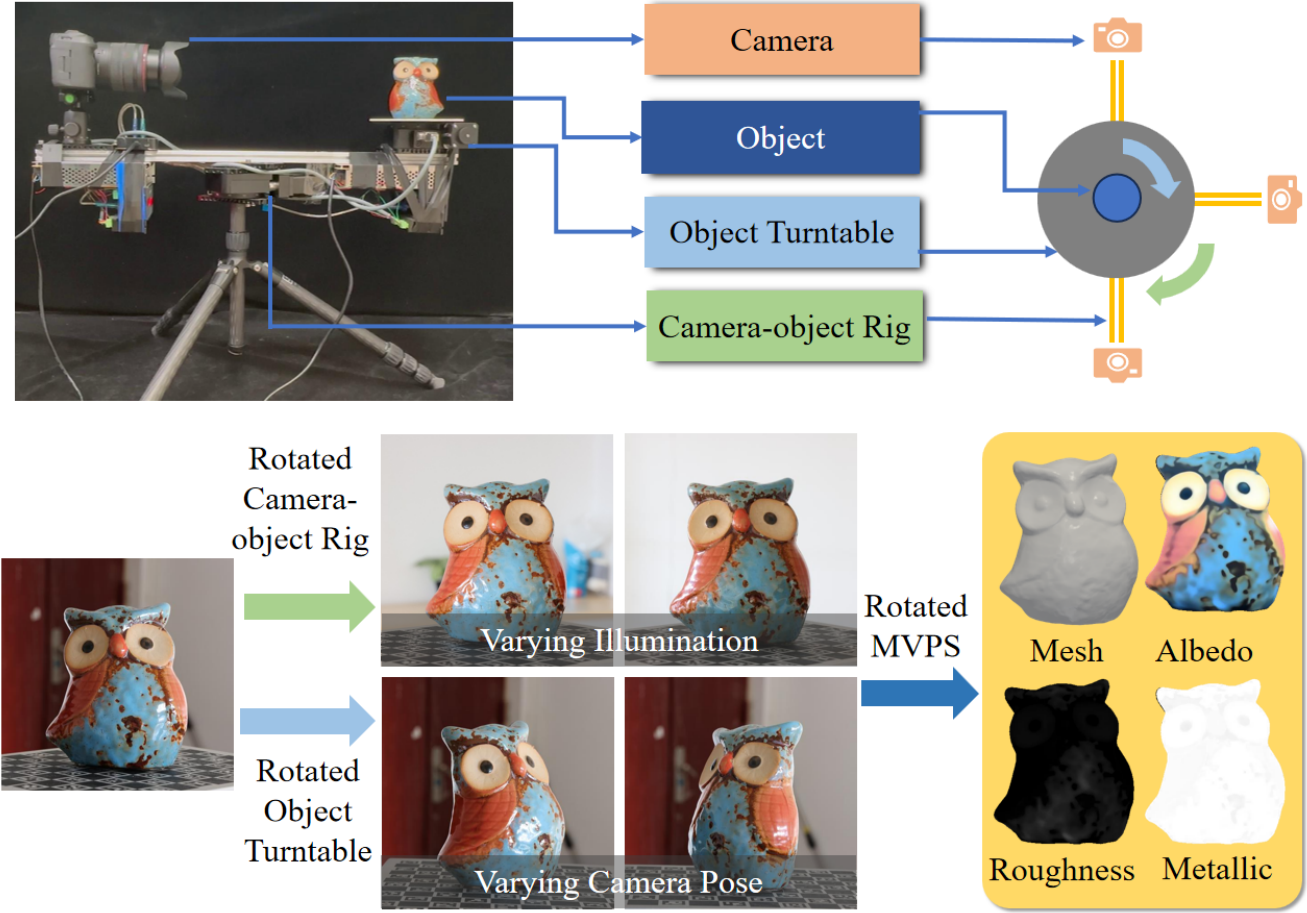}
          \end{overpic}
          \caption{Target object with the turntable and camera are put onto the camera-object rig, from which images under varying views and lights are captured. Our RotatedMVPS leverages rotated light conditions to reconstruct object geometry and reflectance under natural environment light.}
          \label{fig:teaser}
          
\end{figure}

In this paper, we propose RotatedMVPS, a self-supervised MVPS framework based on neural inverse rendering under rotated environment light. To tackle the challenges of complex environment light, we utilize a practical rotation setup. As illustrated in \fref{fig:teaser}, our setup involves jointly rotating the camera and the object with a rig to capture images under varying illuminations but fixed viewpoints. Additionally, the object is rotated independently on a turntable to capture observations from different views. This setup reduces the impact of varying lighting from different directions on the surface reflections of objects in environments with fixed lighting conditions. Minimizing the coupling between ambient light and material properties enhances the accuracy of object shape representation and reflections.


Furthermore, we incorporate learning-based data priors from off-the-shelf single-view photometric stereo methods into our self-supervised MVPS framework. By combining the strengths of self-supervised MVPS with data-driven single-view photometric stereo priors, we effectively enhance the accuracy of shape and reflectance recovery. 

To summarize, the key contributions of this paper are as follows:
\begin{itemize} \item We propose \textbf{RotatedMVPS}, a novel method for recovering surface shape and reflectance under rotated environment light, which reduces the unknowns introduced by environment light. \item We introduce learning-based single-view photometric stereo priors into a self-supervised MVPS framework, demonstrating their effectiveness in boosting performance. \item We design a novel experimental setup and capture a dataset to validate our approach, showcasing its effectiveness through both synthetic and real-world experiments.\end{itemize}

\section{Related Works}

In this section, we review the existing works on neural 3D reconstruction and multi-view photometric stereo, explaining their limitations and how our approach addresses these challenges.
\begin{table}
          \centering
          \caption{\normalfont Comparison of existing works with our method.}
          \label{tab:comparison}
          \begin{tabular}{ccccc}
          \toprule
          \multirow{3}{*}{\textbf{Method}} & \multirow{3}{*}{\textbf{Setup}}& \multirow{3}{*}{\begin{tabular}{c}
               \textbf{Light} \\ \textbf{setting}
          \end{tabular}} & \textbf{BRDF \& }& \multirow{3}{*}{\begin{tabular}{c}
               \textbf{Shape} \\ \textbf{accuracy}
          \end{tabular}}\\
          &&&\textbf{illumination}&
          \\&&&\textbf{estimation}
          \\
          \midrule
          NeuS~\cite{wang2021neus}& MVS& Env. light & \xmark & Low \\
          NeRO~\cite{liu2023nero} &MVS& Env. light & \cmark & Medium \\
          R-MVPS~\cite{park2016robust}&MVPS&Darkroom&\xmark&Low\\
          B-MVPS~\cite{B-MVPS}&MVPS&Darkroom&\xmark&Low\\
          PS-NeRF~\cite{yang2022ps} & MVPS & Darkroom & \xmark & Medium
          \\
            MVPSNet~\cite{zhao2023mvpsnet} & MVPS & Darkroom &\xmark &
          Medium\\
Rnb-NeuS~\cite{brument2024rnb} &MVPS& Env. light &\xmark& Medium\\
          SuperNormal~\cite{cao2024supernormal} &MVPS& Env. light&\xmark& Medium \\
          Ours &MVPS& Env. light&  \cmark & High\\
          \bottomrule
          \end{tabular}
\end{table}
\subsection{Neural 3D Reconstruction}

Neural 3D reconstruction has gained considerable attention due to its ability to recover 3D shapes from multiple 2D images. Early work such as NeRF~\cite{mildenhall2021nerf} demonstrates the power of implicit neural representations for novel view synthesis. Subsequent works~\cite{wang2021neus, yariv2021volume} aims to enhance the efficiency and accuracy of neural 3D reconstruction by using the signed distance function (SDF).

To address the challenge of reconstructing reflective surfaces, several methods~\cite{zhang2021physg, verbin2022ref} decompose the scene into view-dependent reflective appearances, shape, and albedo to improve novel view synthesis and relighting. NeRO~\cite{liu2023nero} combines the integrated direction encoding (IDE) structure from Ref-NeRF~\cite{verbin2022ref} and uses a two-stage optimization process to enhance reconstruction quality. However, these methods primarily focus on static scenes with fixed light conditions.


\subsection{Multiview Photometric Stereo}

Multi-view photometric stereo (MVPS) is a technique for 3D reconstruction that uses multi-view images under varying light conditions. Traditional methods, such as R-MVPS~\cite{R-MVPS} and B-MVPS~\cite{B-MVPS}, begin by initializing the shape through the structure from motion (SfM) and using normal maps for 3D reconstruction.

Kaya et al.~\cite{kaya2022neural} proposed a method for MVPS based on neural radiance field.
 PS-NeRF~\cite{yang2022ps} integrates normal maps from photometric stereo to estimate the geometry and diffuse and specular components.
 MVPSNet~\cite{zhao2023mvpsnet} introduced a fast, deep learning-based solution that utilizes neighbouring views to infer depth and normal maps on a per-view basis. 
As summarized in \Tref{tab:comparison}, these methods are typically evaluated under controlled conditions, such as darkroom with directional light, which limits their applicability to real-world environments with complex light conditions. Recent work SuperNormal~\cite{cao2024supernormal} and 
Rnb-NeuS~\cite{brument2024rnb}, have utilised the state-of-the-art photometric stereo method SDM-UniPS~\cite{ikehata2023scalable} under natural light conditions, achieving high-quality 3D reconstructions.

In contrast, our method leverages neural networks and photometric stereo techniques, specifically in scenarios where both the object and camera rotation capture equivalent rotated light conditions. This approach more effectively handles complex environment light, resulting in a significant improvement in both the quality and realism of the 3D reconstruction.

\section{Method}
\label{sec:method}

In this section, we present our approach for multi-view 3D reconstruction under rotated light conditions. We begin by introducing the rotated platform setup and the associated data acquisition process. We then describe the architecture of our neural network, detailing how we incorporate geometry features derived from photometric stereo. Combining the above two points, we discuss the reconstruction pipeline and related optimization strategies. Finally, we discuss the BRDF recovery optimization based on the rotating platform.

\subsection{Capture Setup}

As illustrated in \fref{fig:teaser}, our capture system consists of two primary components: a camera-object rig and a turntable for the object. The camera-object rig rotates around the center axis, enabling images to be captured under multiple light directions that are suitable for photometric stereo. The turntable rotates the object to facilitate multi-view acquisition. This dual-rotation design ensures a diverse view of light conditions can be obtained from a static light environment, enriching the data for more robust 3D reconstruction.

For simplicity, we assume that the object's center remains fixed and that the environment light is distant. We denote the captured images as:
\begin{equation}
    \mathbf{I} = \{I_{m}^{n} \mid m \in [1,2,3,\ldots,M], n \in [1,2,3,\ldots,N]\},
\end{equation}
where $I_{m}^{n}$ is the $n$-th image captured at the $m$-th degree of the camera-object rig.

In our unique capture setup, the rotation matrices \(\mathbf{R_a}\) and \(\mathbf{R_b}\) represent the orientations of the camera-object rig and the object turntable, respectively. When the turntable rotates, the camera remains stationary. This configuration can be equivalently interpreted as a scenario where the object is fixed while the camera rotates in the opposite direction of the turntable's motion. Simultaneously, the environment light appears to rotate in the same direction as the turntable.

For a given camera ray \(\mathbf{r} = \{o, d\}\), the transformed camera origin and ray direction under our method are expressed as:
\begin{equation}
o' = \mathbf{R^\top_b}o, \quad d' = \mathbf{R^\top_b} d.\textbf{}
\end{equation}
We use the $\{o', d'\}$ as equivalent rays parameters. When the camera-object rig undergoes rotation, the environment light is also affected. For an incoming light direction \(\omega_i\), its transformed direction is given by:
\begin{equation}
\omega_i' = \mathbf{R_a} \mathbf{R_b} \omega_i.
\label{eq:environment_light}
\end{equation}

\begin{figure}
    \centering
    \includegraphics[width = \linewidth]{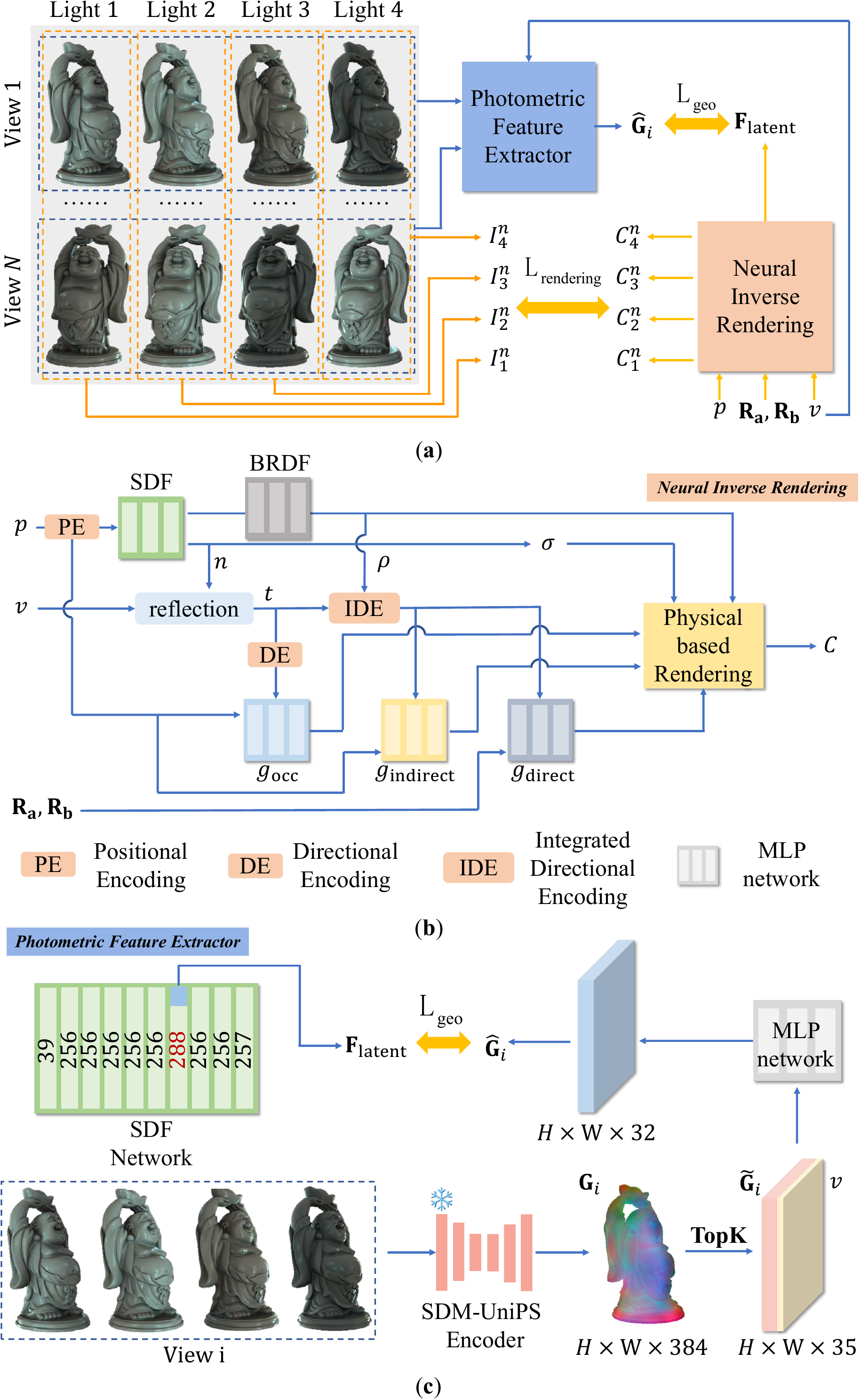}
    \caption{(a) Network architecture overview. We use images under the same light conditions to supervise the radiance field network. For the same view, we use images under different light conditions to extract photometric information using the photometric stereo method. (b) Neural inverse rendering process. (c) The geometry loss $\mathcal{L}_{\text{geo}}$ between the latent feature from the SDF network and the geometric feature from the photometric stereo method. The Latent feature $\mathbf{F}_{latent}$ is the extra feature in the $6$-th layer of the SDF Network.}
    \label{fig:network}
\end{figure}
\subsection{Network Architecture}
\label{sec:network}
~\fref{fig:network} (a) provides an overview of Our network, which is composed of two main parts: a neural inverse rendering module and a photometric feature extractor. 

\paragraph{Neural inverse rendering module}  
We build upon NeRO~\cite{liu2023nero}, a state-of-the-art neural 3D reconstruction framework that represents object surfaces using a signed distance field (SDF) parameterized by a multi-layer perceptron (MLP). For a given camera ray $\mathbf{r} = \{o,d\}$, where $o$ is the camera origin and $d$ is the viewing direction, the network outputs radiance $\mathbf{c}$ and weight $w$ at a sampled point $p$ along the ray. The final rendered color is:
\begin{equation}
    \textbf{C} = \sum_{i} w_i \textbf{c}_i.
\end{equation}

We use the same MLP-based architecture as NeRO~\cite{liu2023nero} to predict BRDF parameters, shown in \fref{fig:network} (b). Let $\omega_o = -d$ be the outgoing view direction and $\omega_i$ be the input light direction on the upper half sphere $\Omega$. The output radiance $\textbf{c}$ is decomposed into diffuse and specular components:
\begin{align}
    \textbf{c} &= \textbf{c}_{\text{diffuse}}+\textbf{c}_{\text{specular}},\\
    \textbf{c}_{\text{diffuse}} &= \frac{\mathbf{a}}{\pi}(1-m)\int_\Omega L(\omega_i)(\omega_i \cdot \mathbf{n})\,d\omega_i,\\
    \textbf{c}_{\text{specular}} &= \int_\Omega L(\omega_i)D(\rho,\mathbf{t})\,d\omega_i \cdot \int_\Omega \frac{DFG}{4(\omega_o \cdot \mathbf{n})}\,d\omega_i,
\end{align}
where $\mathbf{n}$ is the surface normal, $\mathbf{t}$ is the reflection direction, and $L(\omega_i)$ represents the incoming light. Following the micro-facet BRDF model, $D$ is the normal distribution function, $F$ is the Fresnel term, and $G$ is the geometry term. All of them are determined by the albedo $\mathbf{a}$, roughness $\rho$, and metallic $m$.

To separate direct and indirect specular contributions, we follow NeRO~\cite{liu2023nero}:
\begin{equation}
    \begin{aligned}
        \text{L}_{\text{specular}} = [1 - s(\mathbf{t})] g_{\text{direct}}\left(\int_\Omega SH(\mathbf{R_a} \mathbf{R_b} \omega_i)D(\rho,\mathbf{t})\,d\omega_i\right) \\
        + s(\mathbf{t}) g_{\text{indirect}}\left(\int_\Omega SH(\omega_i)D(\rho,\mathbf{t})\,d\omega_i, \mathbf{p}\right),
    \end{aligned}
\end{equation}
where $\text{L}_{\text{specular}}$ is the specular light approximation and $s(\mathbf{t})$ is the occlusion probability, $g_{\text{direct}}$ and $g_{\text{indirect}}$ represent direct and indirect light, respectively, and $SH$ is the spherical harmonics basis function. It is important to highlight the introduction of a rotation-related term, $\mathbf{R_a} \mathbf{R_b}$, as presented in Equation (\ref{eq:environment_light}). This term is essential for accurately estimating environment light in the context of a rotating camera-object rig and turntable setup.


\paragraph{Photometric Feature Extractor}  
Our photometric feature extractor is inspired by the encoder from SDM-UniPS~\cite{ikehata2023scalable}, a transformer-based network that estimates surface normals and BRDF parameters from photometric stereo images. For the $i$-th input view, the SDM-UniPS encoder produces a geometry feature map $\mathbf{G}_{i}$ of dimension $H \times W \times 384$. We compute the variance across feature dimensions and select the top $32$ features with the highest variance:
\begin{equation}
    \widetilde{\mathbf{G}}_{i} = \text{TopK}(\text{Var}(\mathbf{G}_{i}), 32).
\end{equation}

These features are view-dependent. To transform them into world coordinates, we concatenate the view direction $\mathbf{v}_{i}$ and feed them into an MLP:
\begin{equation}
    \hat{\mathbf{G}}_{i} = \text{MLP}(\widetilde{\mathbf{G}}_{i}, \mathbf{v}_{i}),
\end{equation}
resulting in a $H \times W \times 32$ geometry feature map in the world space. As shown in \fref{fig:network} (c), we use these photometric data prior to supervise the $\textbf{F}_{latent}$ in the SDF network, thereby providing additional geometric constraints from the photometric stereo.

\subsection{Training}

Our training objective is a weighted sum of multiple losses:
\begin{equation}
    \mathcal{L} = \mathcal{L}_{\text{rendering}} + \lambda_{\text{eikonal}}\mathcal{L}_{\text{eikonal}} 
    + \lambda_{\text{occ}}\mathcal{L}_{\text{occ}} + \lambda_{\text{geo}}\mathcal{L}_{\text{geo}},
\end{equation}
where we set $\lambda_{\text{eikonal}} =0.1$, 
 $\lambda_{\text{occ}} =1.0 $, $\lambda_{\text{geo}} = 1.0$ in all experiments.

\paragraph{Rendering Loss}
We measure the discrepancy between the rendered and ground-truth images using the same loss as NeRO~\cite{liu2023nero}:
\begin{equation}
    \mathcal{L}_{\text{rendering}} = \sqrt{\sum (I - C)^2}.
\end{equation}

\paragraph{Eikonal Loss}
The eikonal loss~\cite{gropp2020implicit} regularizes the SDF gradients to enforce a unit-norm constraint on surface normals:
\begin{equation}
    \mathcal{L}_{\text{eikonal}} = \sum (||\mathbf{n}|| - 1)^2.
\end{equation}

\paragraph{Occlusion Loss}
Following NeRO~\cite{liu2023nero}, we predict an occlusion probability $s$ from an MLP $g_{occ}$ and supervise it using $s_{march}$ derived from the SDF. The occlusion loss is:
\begin{equation}
    \mathcal{L}_{\text{occ}} = ||s - s_{march}||_1.
\end{equation}

\paragraph{Geometry Loss}
We supervise the part of layer in the SDF network using the geometry feature map $\hat{\mathbf{G}}_{i}$:
\begin{equation}
    \mathcal{L}_{\text{geo}} = ||\hat{\mathbf{G}}_{i} - \mathbf{F}_{latent}||_2.
\end{equation}

To further refine the BRDF parameters, we adopt the same optimization strategy as NeRO~\cite{liu2023nero}, employing Monte Carlo sampling. The detailed BRDF loss formulation can be found in the original paper. During this secondary optimization stage, the object shape is fixed, and thus we no longer use the geometry feature map $\hat{\mathbf{G}}_{i}$.

\section{Experiments}
This section introduces the dataset used in our rotated platform setup and compares our method with other MVPS and MVS approaches in terms of shape recovery and relighting results. Additionally, we conduct an ablation study to evaluate the effectiveness of rotated setup and data priors.

\paragraph{Dataset}
We present two datasets for evaluation: a synthetic dataset and a real-world dataset. The synthetic dataset includes three objects rendered using Blender under the conditions of our rotated setup. For the real-world dataset, we employ a Canon EOS5 camera mounted on the rotation platform to capture images of objects at each rotation step. Specifically, the Camera-object rig rotates by $90^\circ$ per step, while the turntable rotates by $14.4^\circ$ per step, resulting in $4$ distinct light conditions and $25$ unique views per object. The resolution of the captured images is $512 \times 512$. To ensure a fair comparison with MVS methods such as NeRO~\cite{liu2023nero} and VMINer~\cite{vminer}, we render images in 100 different views, maintaining the same number of input images for both MVS and MVPS methods. However, for the real-world dataset, we only compare against MVPS methods because the images captured under the turntable setup do not correspond to a fixed environment light condition.

\paragraph{Baseline}
We compare our method with three MVS methods, NeuS~\cite{wang2021neus} , NeRO~\cite{liu2023nero} and VMINer~\cite{vminer}, and two MVPS methods, SuperNormal~\cite{cao2024supernormal} and Rnb-NeuS~\cite{brument2024rnb}.

\paragraph{Evaluation metrics}
We use Chamfer Distance (CD) to measure the accuracy of 3D reconstruction. We also compare the re-rendered images with ground-truth images to evaluate rendering quality using the Structural Similarity Index (SSIM) and Peak Signal-to-Noise Ratio (PSNR).

\begin{figure}
          \centering
          \begin{overpic}[width=1\linewidth]{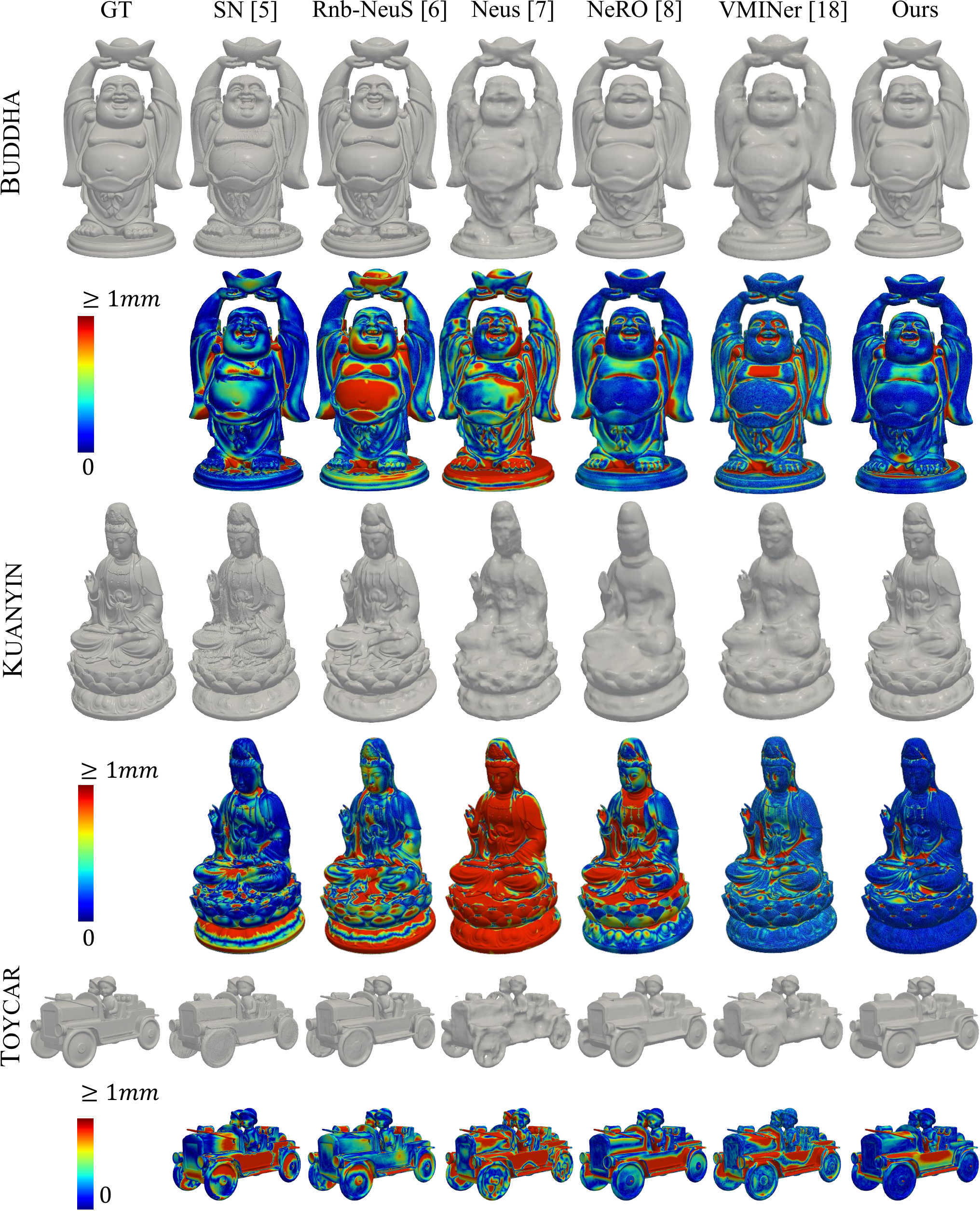}
          \end{overpic}
          \caption{Qualitative comparison of recovered shapes of BUDDHA, KWANYIN, TOYCAR and their CD error maps.}
          \label{fig:synthetic}
        \vspace{-0em}
\end{figure}

\paragraph{Implementation details}
We implement our method in PyTorch 1.10.0 and train the model on a single NVIDIA RTX 4090 GPU. We use the Adam optimizer with a learning rate of $1e-4$ and train the model for $300,000$ epochs. 

\subsection{Reconstruction results on synthetic dataset}
We first evaluate the performance of our method on the synthetic dataset. The quantitative results are shown in \Tref{tab:synthetic}. Our approach surpasses all baseline methods in terms of Chamfer Distance (CD), indicating superior 3D reconstruction accuracy. The qualitative results are shown in \fref{fig:synthetic}, where our method demonstrates enhanced object surface detail compared to MVS techniques. For example, the face of the {\sc{Kwanyin}} is more accurate than \neus~, \nero~and \vminer. This is because MVS methods focus on whole-shape reconstruction, and the object's details may be influenced by light variance. Compared with MVPS methods, although they seem to have more surface detail, the accuracy of the shape is not as good as our method. This is because MVPS methods rely on the normal map from photometric stereo, which may be influenced by reflective surfaces when light changes. Our method combines the advantages of MVS and MVPS methods, focusing on whole shape reconstruction and utilizing geometry features from photometric stereo to improve object surface detail. 
\begin{table}[t]
    \centering
    \caption{Comparison of shape recoveries on synthetic datasets, measured by CD in millimeters (mm). }\label{B}
    \renewcommand{\arraystretch}{1.5}
        
    \begin{tabular}{@{}l|ccc}
        \hline
        Methods   & {\sc{Kwanyin}} & {\sc{Buddha}} & {\sc{Toycar}}  \\
        \hline
        \supernormal & 1.3329 & 0.7025  & 1.5161     \\
        \rnbneus     & 1.9111 & 1.4971  &1.7373	     \\
        \neus    & 1.9346 & 1.0945  &1.4994	     \\ 
        \nero        & 1.6532 & 0.4522	&1.4658	     \\
        \vminer      & 1.7812 & 0.5280  & 1.3232	     \\
        \hline  
        Ours         & \textbf{1.0157} & \textbf{0.3651}	& \textbf{1.2403}      \\ 
        \hline 
    \end{tabular}
    \label{tab:synthetic}
\end{table}
\begin{figure}
          \centering
          \begin{overpic}[width=1\linewidth]{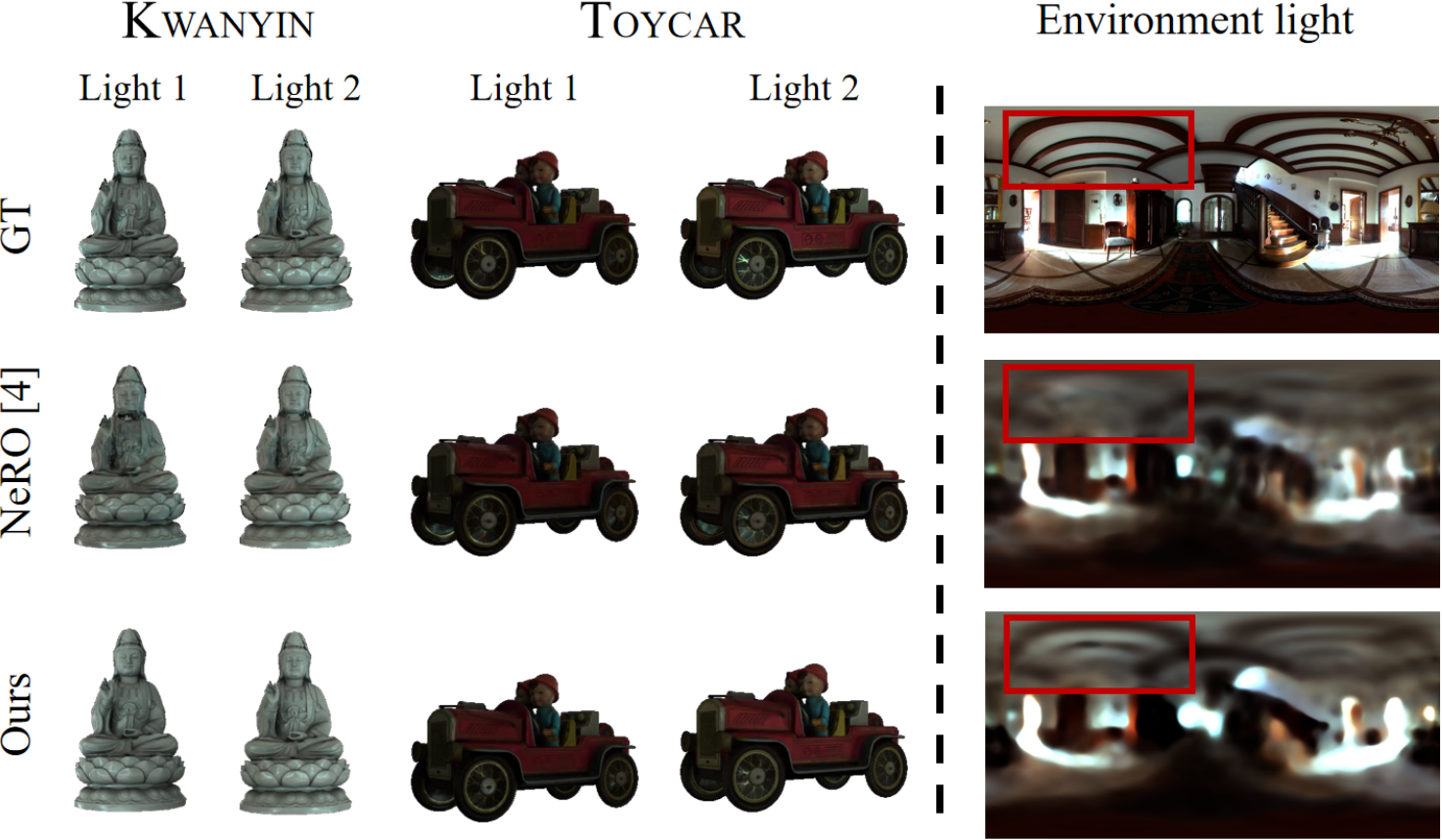}
          \end{overpic}
          \caption{(Left) Qualitative comparison of relighting result by using different environment light. (Right) Recovered environment light.}
          \label{fig:brdf}
          \vspace{-1em}
\end{figure}

Our method demonstrates superior inverse rendering performance, as evidenced by the quantitative metrics presented in \Tref{tab:brdf}. Our re-rendered images achieve higher PSNR and SSIM values, indicating improved fidelity to the ground truth images. \fref{fig:brdf} further illustrates that our re-rendered outputs closely resemble the ground truth, surpassing the quality achieved by \nero. Relighting results show that our method has more accurate and convincing texture results. Notably, the relight quality on the neck of the {\sc{Kwanyin}} model is enhanced in our results, which is attributable to our more precise shape reconstruction. Additionally, our environment light predictions capture finer details compared to \nero, as highlighted by the red box in the room's ceiling area.

\begin{table}[t]
    \centering
    \caption{Comparison of reflectance recoveries on our synthetic dataset based on image relighting.}
    \renewcommand{\arraystretch}{1.5}
    \begin{tabular}{@{}l|cc|cc}
        \hline
        \multirow{2.4}{*}{Methods} &  \multicolumn{2}{c|}{PSNR $\uparrow$} &\multicolumn{2}{c}{SSIM $\uparrow$} \\ 
        ~      & {\sc{Kwanyin}}  & {\sc{Toycar}} & {\sc{Kwanyin}}  & {\sc{Toycar}}  \\
        \hline
        \nero     & 35,26 & 38.78   &0.9850	    &0.9819    \\  
        Ours      & \textbf{37.73} & \textbf{39.86}	&  \textbf{0.9894} 	&  \textbf{0.9822}    \\ 
        \hline
    \end{tabular}
    \label{tab:brdf}
\end{table}

\subsection{Reconstruction results on real-world dataset}
\begin{figure}
          \centering
          \begin{overpic}[width=1\linewidth]{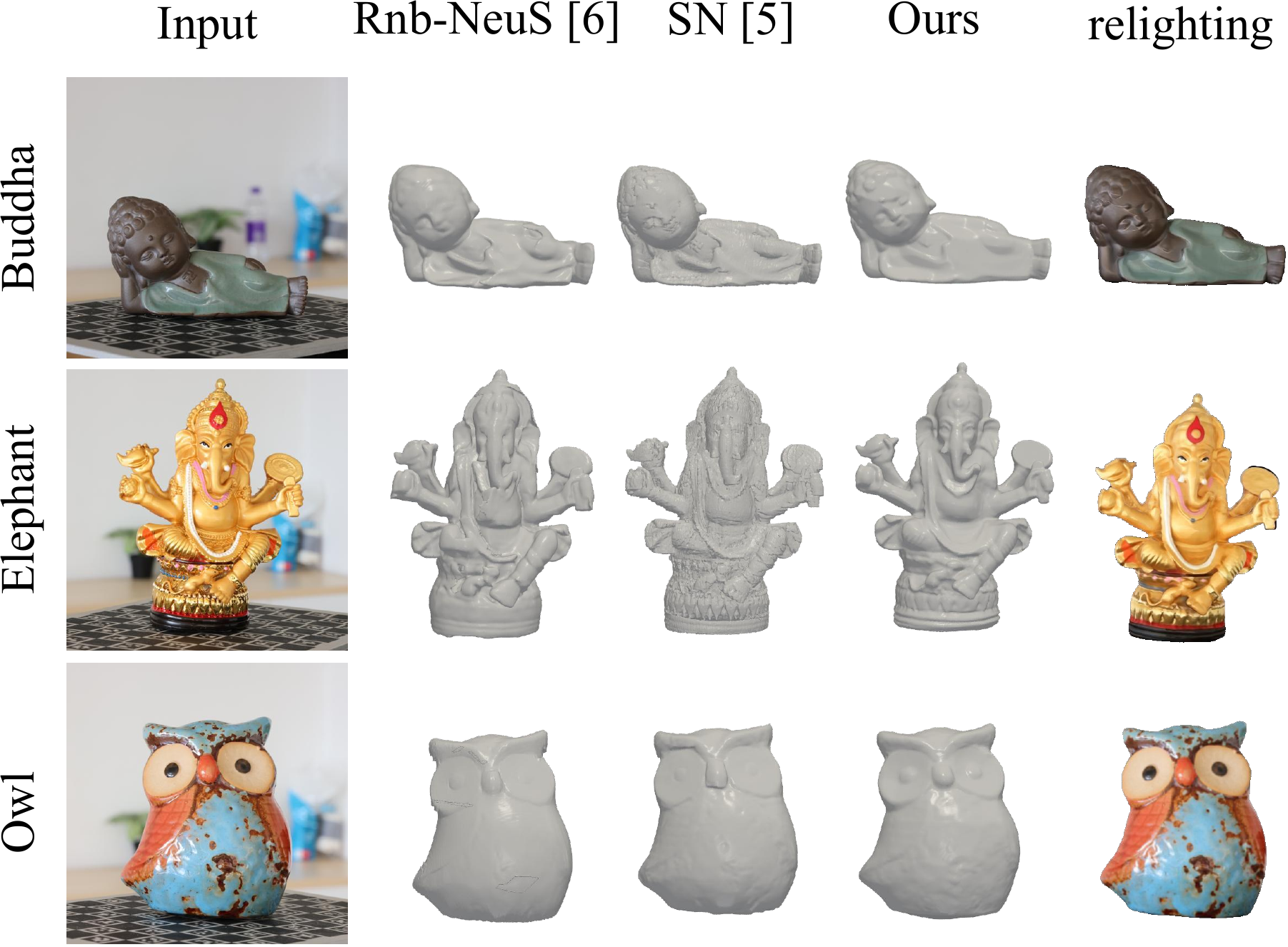}

          \end{overpic}
          \caption{Qualitative comparison on shape recovery and image relighting results on our real-captured data.}
          \label{fig:real_world}  \vspace{-1em}
\end{figure}

We evaluated our method's performance on a real-world dataset, with qualitative results presented in \fref{fig:real_world}. Our approach effectively mitigates the propagation of photometric stereo errors into shape reconstruction, a challenge observed in methods like Rnb-Neus\cite{brument2024rnb} and SuperNormal~\cite{cao2024supernormal}. These methods introduce inaccuracies in detailed regions of the model due to their reliance on photometric stereo data. In the reconstructions of the Buddha and Elephant models, previous MVPS methods exhibit sharp, unrealistic discontinuities and gaps. In contrast, our method achieves reconstructions with a high degree of integrity, free from such anomalies, even in intricate details. This improvement is attributed to our MVS and MVPS techniques, which enhance overall shape accuracy and surface detail.

\begin{figure}
          \centering
          \begin{overpic}[width=1\linewidth]{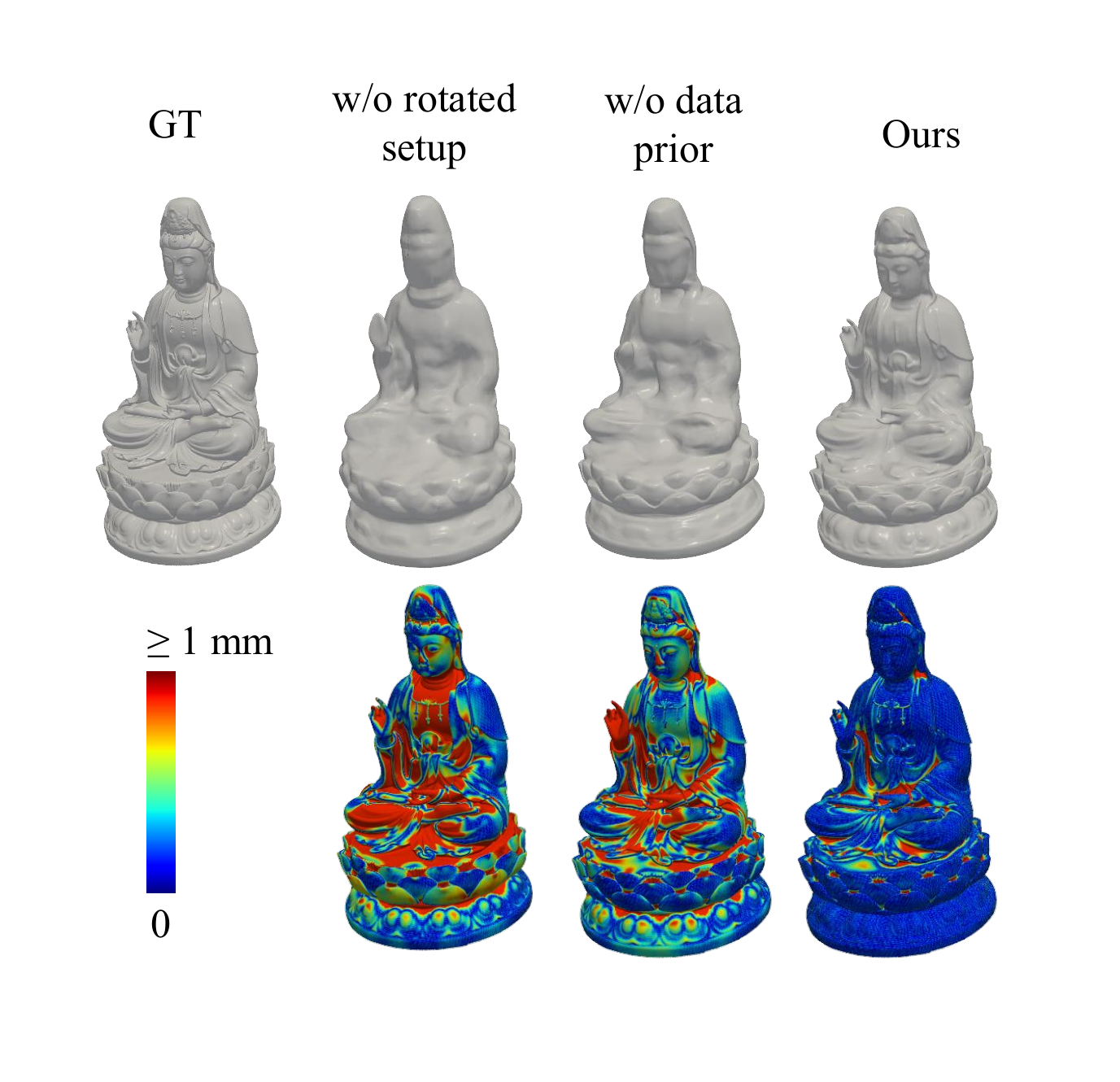}

          \end{overpic}
          \caption{Ablation study on {\sc{Kwanyin}}. The top and bottom rows show estimated shape and error maps of CD.}
          \label{fig:ablation}
          
\end{figure}

\begin{table}[t]
    \centering
    \caption{Ablation study of the rotated setup and data prior measured by CD in millimeters (mm).}
    \renewcommand{\arraystretch}{1.5}
    \begin{tabular}{@{}l|ccc|c}
        \hline
        Methods   & {\sc{Kwanyin}} & {\sc{Buddha}} & {\sc{Toycar}} &{Average} \\
        \hline
       
        w/o rotated setup  & 1.6532 & 0.4522    &1.4658	  &1.1904  \\  
        w/o data prior     & 1.2942 & 0.5602    & 1.4244  &1.0930  \\
        Ours               & \textbf{1.0157} & \textbf{0.3651}	& \textbf{1.2403}  & \textbf{0.8738}   \\ 
        \hline
    \end{tabular}
    \label{tab:ablation}
    \vspace{-0em}
\end{table}

\subsection{Ablation study}
We conducted an ablation study to assess the impact of incorporating photometric stereo-derived geometric features into our 3D reconstruction method. The quantitative outcomes are presented in  \Tref{tab:ablation}, and the visual comparisons are illustrated in \fref{fig:ablation}. Initially, we trained the model without the rotated-platform setup, corresponding to the baseline results akin to NeRO~\cite{liu2023nero}. Subsequently, we trained the model without integrating the MVPS geometric features. The findings indicate that incorporating these geometric features significantly enhances both the accuracy of shape reconstruction and the fidelity of object surface details, surpassing the performance achieved by relying solely on normal maps from photometric stereo.
Moreover, the results demonstrate that even in the absence of geometric features, our specific imaging setup contributes to an overall improvement in shape reconstruction quality. The ablation study suggests that our method's design effectively leverages the advantages of both MVS and MVPS techniques, leading to more precise and detailed 3D reconstructions.
\section{Conclusion}
In this paper, we propose a multi-view photometric stereo method under rotated environment light. By introducing a practical rotated environment light setup, we reduce the coupling of environment light and surface reflectance and increase the recovery accuracy of both surface shape and reflectance. Additionally, we leverage data priors from off-the-shelf learning-based single-view photometric stereo methods, integrating them into our self-supervised MVPS framework. This combination of techniques significantly enhances the performance of shape and reflectance recovery. Through both synthetic and real-world experiments, we demonstrate the effectiveness of our proposed approach, offering a promising solution for inverse rendering tasks in natural light conditions.
\section{Acknowledgment}
This work was supported by National Natural Science Foundation of China (Grant No. 62472044, U24B20155), Hebei Natural Science Foundation Project No. F2024502017, Beijing-Tianjin-Hebei Basic Research Funding Program No. 242Q0101Z. We thank openbayes.com for providing computing resource.

\bibliographystyle{IEEEbib}
\bibliography{icme2025references}

\end{document}